\documentclass[11pt,letterpaper]{article}
\usepackage{emnlp2017}
\usepackage{times}
\usepackage{latexsym}
\usepackage{xcolor}
\usepackage{graphicx}
\usepackage{fancyvrb}
\usepackage{url}
\usepackage{paralist}
\definecolor{dgreen}{rgb}{0.0,0.7,0.0}
\definecolor{lblue}{rgb}{0.2,0.5,1}
\newcommand{\f}[1]{\textcolor{lblue}{~~~~~#1}}

\newcommand{\m}[1]{\textcolor{dgreen}{\em ~~~~~#1}}
\newif\ifarxiv
\arxivtrue

\newenvironment{MyColorPar}[1]{%
    \leavevmode\color{#1}\ignorespaces%
}{%
}%
\definecolor{dblue}{rgb}{0,0,0.25}

\emnlpfinalcopy

\def\emnlppaperid{***}

\title{ParlAI: A Dialog Research Software Platform}

\author{Alexander H. Miller, Will Feng, Adam Fisch,    Jiasen Lu,\\
{\bf Dhruv Batra, Antoine Bordes, Devi Parikh  \and Jason Weston}  \\
 { Facebook AI Research}}

\date{}

\begin{document}

\maketitle

\begin{abstract}
We introduce ParlAI (pronounced ``par-lay''), an open-source software platform for dialog research implemented in Python, available at {\small\url{http://parl.ai}}.
Its goal is to provide a unified framework for sharing, training and testing dialog models; integration of Amazon Mechanical Turk for data collection, human evaluation, and online/reinforcement learning; and a repository of machine learning models for comparing with others' models, and improving upon existing architectures.
Over 20 tasks are supported in the first release, including popular datasets such as SQuAD, bAbI tasks, MCTest, WikiQA, 
QACNN, QADailyMail, CBT, 
bAbI Dialog, Ubuntu, OpenSubtitles 
 and VQA. 
Several models are integrated, including neural models such as memory networks, seq2seq and attentive LSTMs.
\end{abstract}

\section{Introduction}

The purpose of language is to accomplish communication goals, 
%
which typically involve a dialog between two or more communicators \cite{crystal2004cambridge}.
Hence, trying to solve dialog is a fundamental goal for researchers in the NLP community.
From a machine learning perspective,  building a learning agent capable of dialog
is also fundamental for various reasons, chiefly that the solution involves achieving
most of the subgoals of the field, and in many cases those subtasks are directly impactful to the task.
%

\begin{figure}[t!]
\begin{tiny}
\includegraphics[width=0.5\textwidth]{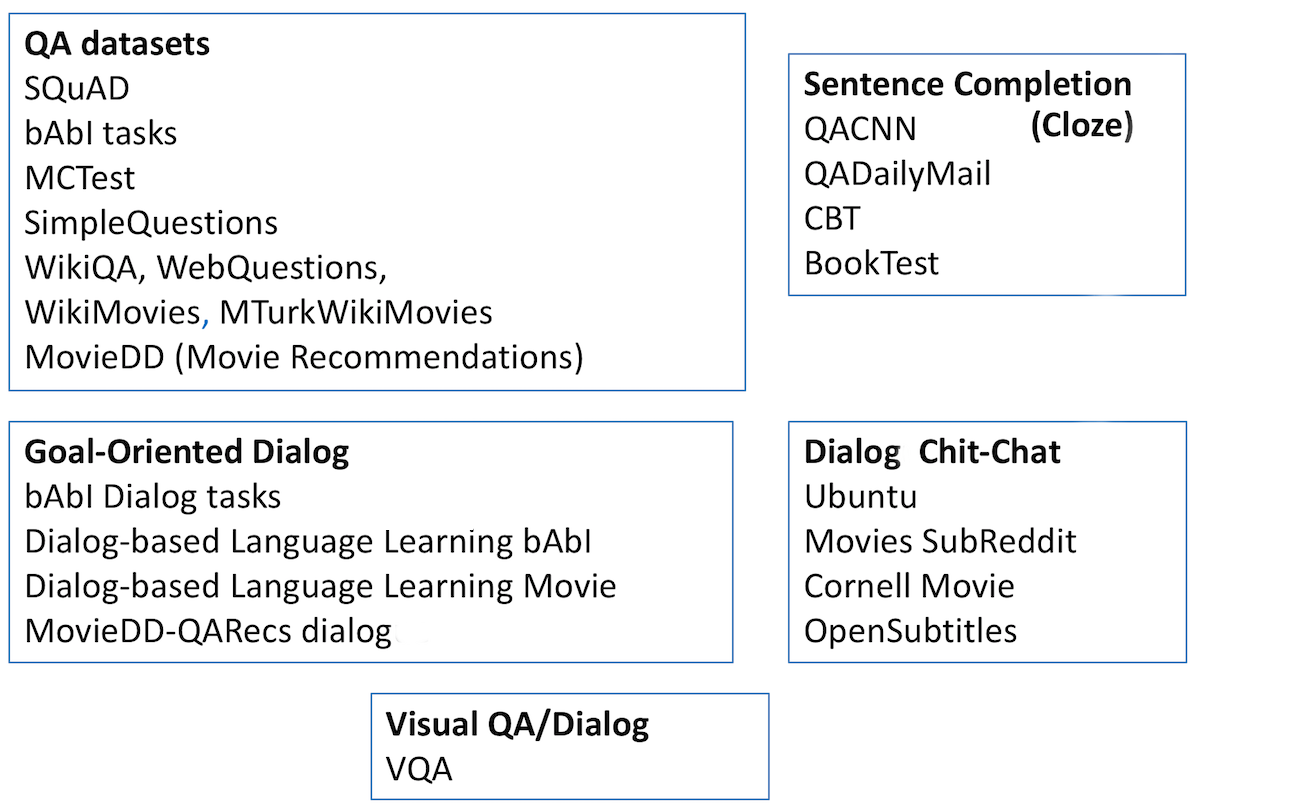}
\end{tiny}
\vspace{-8mm}
\caption{
The tasks in the first release of ParlAI.
\label{fig:tasks}
}
\vspace{2mm}
\center
\begin{tiny}
\includegraphics[width=0.44\textwidth]{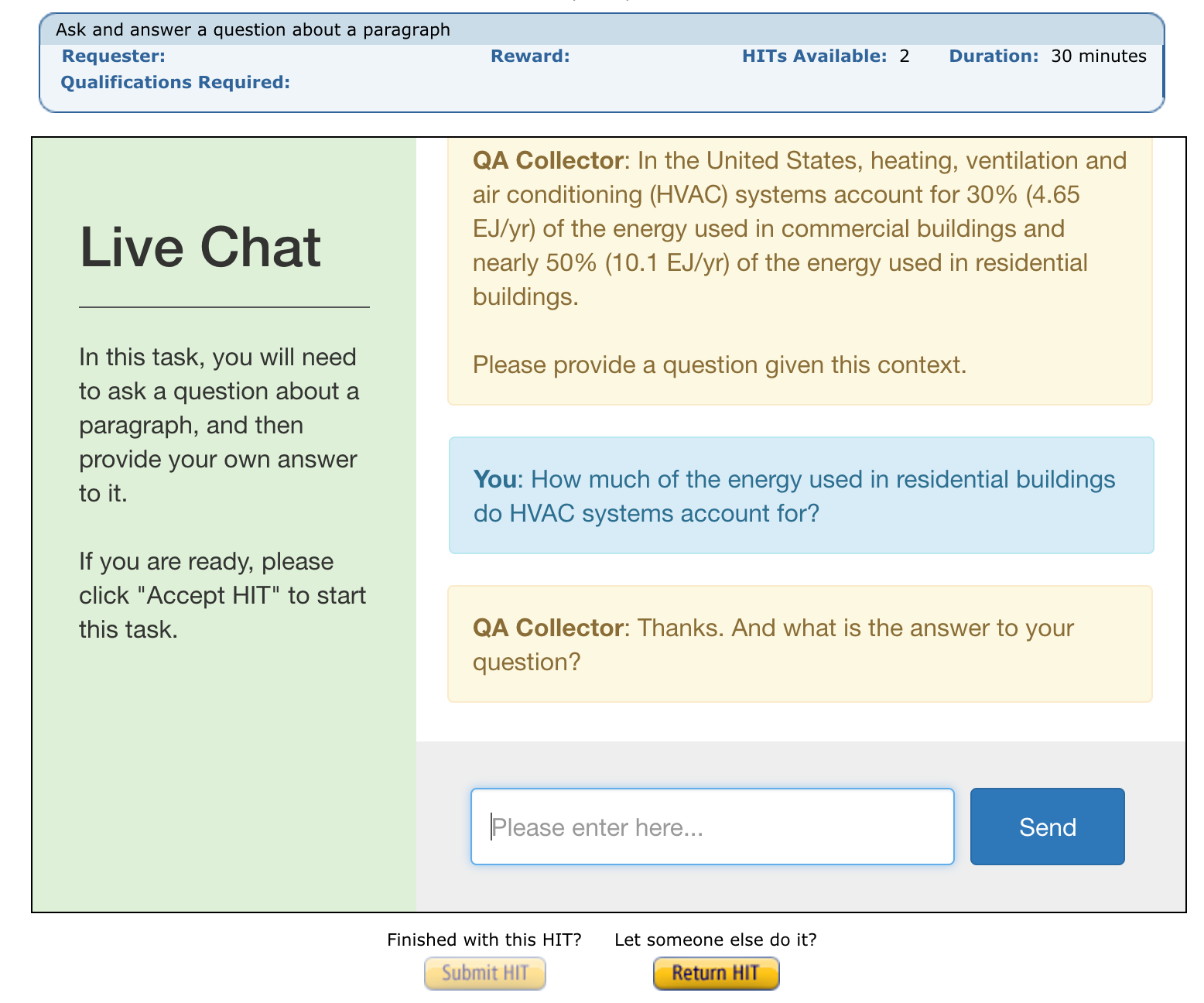}
\end{tiny}
\caption{
MTurk Live Chat for collecting QA datasets in ParlAI.
\label{fig:mturk}
}
\vspace{-4mm}
\end{figure}

On the one hand dialog can be seen as a single task (learning how to talk) and on the other hand as thousands of related tasks that require different skills, all using the same input and output format.
The task of booking a restaurant, chatting about sports or the news, or answering factual or perceptually-grounded 
questions all fall under dialog. 
Hence, methods that perform task transfer appear useful for the end-goal.
Memory, logical and commonsense reasoning, planning, learning from interaction, 
learning compositionality  and other AI subgoals also have clear roles in dialog.


However, to pursue these research goals, software tools should unify 
the different dialog sub-tasks
and the agents that can learn from them. 
Working on individual datasets can lead to siloed research,
where the overfitting to specific qualities of a dataset 
do not generalize to solving other tasks.
\ifarxiv
{For example, methods that do not generalize beyond WebQuestions \cite{berant2013semantic} because they specialize on knowledge bases only,  SQuAD  \cite{rajpurkar2016squad}  because they predict start and end context indices (see Sec. \ref{sec:exp}), or  bAbI \cite{weston2015towards}   because they use supporting facts  or make use of its simulated nature.
}
\fi

In this paper we present a software platform, ParlAI (pronounced ``par-lay''), that provides researchers
a unified framework for training and testing dialog models, especially multitask training or evaluation over many tasks at once, 
as well as seamless integration with Amazon Mechanical Turk.
Over 20 tasks are supported in the first release, including many popular datasets, see Fig.~\ref{fig:tasks}.  Included are examples of training neural models with PyTorch 
and  Lua Torch\footnote{\tiny\url{http://pytorch.org/} and \tiny\url{http://torch.ch/}}.
Using Theano\footnote{\tiny\url{http://deeplearning.net/software/theano/}} or Tensorflow\footnote{\tiny\url{https://www.tensorflow.org/}}
 instead is also straightforward.

The overarching goal of ParlAI is to build a community-based platform for easy access to both tasks and learning algorithms that perform well on them, in order to push the field forward.
This paper describes our goals in detail, and gives a technical overview of the platform. 


%
%
%
%
%
%

\section{Goals}

The goals of ParlAI are as follows:

\vspace{-5pt}
\paragraph{A unified framework for development of dialog models.}
ParlAI aims to unify dialog dataset input formats
fed to machine learning agents to a {\em single format}, and to standardize 
 evaluation frameworks and metrics as much as possible. 
Researchers can submit their new tasks and their agent training code to the repository to share with others in order to aid reproducibility, and to better enable follow-on research.

\vspace{-5pt}
\paragraph{General dialog involving many different skills.}
ParlAI contains a seamless combination of real and simulated language datasets, and encourages multitask model development \& evaluation by making multitask models as easy to build as single task ones.  
This should reduce overfitting of model design to specific datasets and encourage models that perform task transfer, an important prerequisite for a general dialog agent.

\vspace{-5pt}
\paragraph{Real dialog with people.}
ParlAI allows collecting, training and evaluating on live dialog with humans via Amazon Mechanical Turk by 
making it easy to connect Turkers with a dialog agent, see Fig.~\ref{fig:mturk}.
This also enables comparison of Turk experiments across different research groups, which has been historically difficult.

\vspace{-5pt}
\paragraph{Towards a common general dialog model.}
Our aim is to motivate the building of new tasks and agents that move the field towards a working dialog model. Hence, each new task that goes into the repository should build towards that common goal, rather than being seen solely as a piece of independent research. 

\section{General Properties of ParlAI}

ParlAI  consists of a number of tasks and agents that can be used to solve them. All the tasks in ParlAI have a single format (API) which makes applying any agent to any task,
or multiple tasks at once, simple.
The tasks include both fixed supervised/imitation learning datasets (i.e.~conversation logs) and interactive (online or reinforcement learning) tasks, as well as both real language and simulated tasks, which can all be seamlessly trained on.
ParlAI also supports other media, e.g.~images as well as text for 
\ifarxiv
visual question answering~\cite{antol2015vqa}  or visually grounded dialog~\cite{das2017learning}. 
\fi
ParlAI automatically downloads tasks and datasets the first time they are used.
One or more Mechanical Turkers can be embedded inside an environment (task) to collect data, train or evaluate learning agents.

Examples are included in the first release of training with PyTorch and Lua Torch.
ParlAI uses ZeroMQ to talk to languages other than Python (such as Lua Torch).
Both batch training and hogwild training of models are supported and built into the code.
An example main for training an agent is given in Fig.~\ref{fig:example_main}.

\section{Worlds, Agents and Teachers}

The main concepts (classes) in ParlAI are worlds, agents and teachers:

\begin{compactitem}
\item {\bf world} -- the environment. This can vary from being very simple, e.g.~just two agents conversing, to much more complex, e.g.~multiple agents in an interactive environment.
\item {\bf agent} -- an agent that can act (especially, speak) in the world. An agent is either a learner (i.e.~a machine learned system), a hard-coded bot such as one designed to interact with learners, or a human (e.g.~a Turker).
\item {\bf teacher} -- a type of agent that talks to the learner in order to teach it, e.g.~implements one of the tasks in Fig.~\ref{fig:tasks}.
\end{compactitem}

After defining a world and the agents in it, a main loop can be run for training, testing or displaying, which calls the function world.parley() to run one time step of the world. Example code to display data is given in Fig.~\ref{fig:example_main}, and the output of running it is in Fig.~\ref{fig:example_output}.


\begin{figure}[t]
\begin{tiny}
\begin{MyColorPar}{blue}
\begin{Verbatim}[frame=single]
teacher = SquadTeacher(opt)
agent = MyAgent(opt)
world = World(opt, [teacher, agent])
for i in range(num_exs):
  world.parley()
  print(world.display())
\end{Verbatim}
\end{MyColorPar}
\end{tiny}
\vspace{-5mm}
\begin{tiny}
\begin{MyColorPar}{dgreen}
\begin{Verbatim}[frame=single]
def parley(self): 
  for agent in self.agents:            
    act = agent.act()
    for other_agent in self.agents:
      if other_agent != agent:
        other_agent.observe(act)
\end{Verbatim}
\end{MyColorPar}
\end{tiny}
\caption{
ParlAI main for displaying data (top) and the code for the world.parley call (bottom).
\label{fig:example_main}
}
\end{figure}

\section{Actions and Observations} \label{sec:oad}

All agents (including teachers) speak to each other in a single common format -- the observation/action object (a python dict), see Fig.~\ref{fig:oad}.  
It is used to pass text, labels and rewards between agents. The same object type is used for both talking (acting) and listening (observing), but with different values in the fields.  Hence, the object is returned from agent.act() and passed in to agent.observe(), see Fig.~\ref{fig:example_main}.

The fields of the message are as follows: 
\begin{compactitem}
\item {\em text:} a speech act.
\item {\em id:} the speaker's identity.
\item {\em reward:} a real-valued reward assigned to the receiver of the message.
\item {\em episode\_done:}  indicating the end of a dialog.
\end{compactitem}

For supervised datasets, there are some additional fields that can be used: 
\begin{compactitem}
\item {\em label:}  a set of answers the speaker is expecting to receive in reply, e.g.~for QA datasets the right answers to a question.
\item {\em label\_candidates:} a set of possible ways to respond supplied by a teacher, e.g.~for multiple choice datasets or ranking tasks.
\item {\em text\_candidates:} ranked candidate predictions from a learner. Used to evaluate ranking metrics, rather than just evaluate the single response in the {\em text} field.
\item {\em metrics:} A teacher can communicate to a learning agent metrics on its performance.
\end{compactitem}

Finally other media can also be supported with additional fields:
\begin{compactitem}
\item {\em image:} an image, e.g.~for Visual Question Answering or Visual Dialog datasets.
\end{compactitem}
As the dict is extensible, we can add more fields over time,
e.g.~for audio and other sensory data, as well as actions other than speech acts.

\begin{figure}
\begin{tiny}
\begin{MyColorPar}{dblue}
\begin{Verbatim}[frame=single]
python examples/display_data.py -t babi 

[babi:Task1k:4]: The office is north of the kitchen.
The bathroom is north of the office.
What is north of the kitchen?
[cands: office|garden|hallway|bedroom|kitchen|bathroom]
   [RepeatLabelAgent]: office
- - - - - - - - - - - - - - - - - - - - -
~~
[babi:Task1k:2]: Daniel went to the kitchen.
Daniel grabbed the football there.
Mary took the milk there.
Mary journeyed to the office.
Where is the milk?
[cands: office|garden|hallway|bedroom|kitchen|bathroom]
   [RepeatLabelAgent]: office
\end{Verbatim}
\end{MyColorPar}
\end{tiny}
\caption{
Example output to display data of a given task
(see Fig.~\ref{fig:example_main} for corresponding code).
\label{fig:example_output}
}
\end{figure}


Each of these fields are technically optional, depending on the dataset, though the {\em text} field will most likely be used in nearly all exchanges.
A typical exchange from a ParlAI training set is shown
in Fig.~\ref{fig:typical}.

\section{Code Structure}

The ParlAI codebase has five main directories:

\begin{compactitem}
\item{\bf core}:  the primary code for the platform.
\item{\bf agents}: contains agents which can interact with the worlds/tasks (e.g.~learning models).
\item{\bf examples}: contains examples of different mains (display data, training and evaluation).
\item{\bf tasks}: contains code for the different tasks available from within ParlAI.
\item{\bf mturk}: contains code for setting up Mechanical Turk and sample MTurk tasks.
\end{compactitem}

\begin{figure}[t]
\begin{small}
\fbox{\begin{minipage}{23em}
\textcolor{red}{\bf Observation/action {\tt dict}}\\
Passed back and forth between agents \& environment.\\
\\
{\em Contains:} \\
\f{.text~~~~~~~~~~~~~~~~}            \m{text of speaker(s)}\\
\f{.id~~~~~~~~~~~~~~~~~~~}              \m{id of speaker(s)}\\
\f{.reward~~~~~~~~~~~}          \m{for reinforcement learning}\\
\f{.episode\_done}    \m{signals end of episode}\\
\\
{\em For supervised dialog datasets:}\\
\f{.label}\\
\f{.label\_candidates}  \m{multiple choice options}\\
\f{.text\_candidates}   \m{~~ranked candidate responses} \\
\f{.metrics}           \m{~~~~~~~~~~~~~~evaluation metrics}\\
\\
{\em Other media:}\\
\f{.image}             \m{~~~~~~~~~~~~~~~~for VQA or Visual Dialog}\\
\end{minipage}}
\end{small}
\caption{The observation/action dict  is the central message passing object in ParlAI:  agents send this message to speak, and receive a message of this form to observe other speakers and the environment.
\label{fig:oad}
}
\end{figure}

\subsection{Core}

The core library contains the following files:

\begin{compactitem}
\item {\tt\small agents.py}: defines the Agent base class for all agents, which implements the observe() and act() methods, the Teacher class which also reports metrics, and MultiTaskTeacher for multitask training.
\item {\tt\small dialog\_teacher.py}: the base teacher class for doing dialog with fixed chat logs.
\item {\tt\small worlds.py}: defines the base World class, DialogPartnerWorld for two speakers,
MultiAgentDialogWorld for more than two,
and two containers that can wrap a chosen environment: BatchWorld for batch training, and HogwildWorld for training across multiple threads.
\ifarxiv
\item {\tt\small dict.py}: code for building language dictionaries.
\item {\tt\small metrics.py}: computes exact match, F1 and ranking metrics for evaluation.
\item {\tt\small params.py}: uses argparse to interpret command line arguments for ParlAI
\fi

\end{compactitem}

\subsection{Agents}

The agents directory contains machine learning agents.
Currently available within this directory:

\begin{compactitem}
\item {\bf drqa:} an attentive LSTM model DrQA \cite{drqa} implemented in PyTorch that has competitive results on SQuAD  \cite{rajpurkar2016squad}  amongst other datasets.
\item {\bf memnn:} code for an end-to-end memory network \cite{memn2n} in Lua Torch.
\ifarxiv
\item {\bf remote\_agent:} basic class for any agent connecting over ZeroMQ.
\fi
\item {\bf seq2seq:} basic GRU sequence to sequence model \cite{sutskever2014sequence}
\item {\bf ir\_baseline:} information retrieval (IR) baseline that scores responses with TFIDF-weighted matching \cite{ritter2011data}.
\ifarxiv
\item {\bf repeat\_label:} basic class for merely repeating all data sent to it (e.g.~for debugging).
\fi
\end{compactitem}


\begin{figure}
\begin{tiny}
\vspace{-4mm}
\begin{MyColorPar}{blue}
\begin{Verbatim}[frame=single]
Teacher: {
    'text': 'Sam went to the kitchen.\n Pat gave Sam the 
milk.\nWhere is the milk?',\\
    'labels': ['kitchen'],
    'label_candidates': ['hallway', 'kitchen', 'bathroom'],
    'episode_done': False
}

Student: {
   'text': 'hallway'
}

Teacher: {
    'text': 'Sam went to the hallway\nPat went to the 
bathroom\nWhere is the milk?',
    'labels': ['hallway'],
    'label_candidates': ['hallway', 'kitchen', 'bathroom'],
    'done': True
}

Student: {
    'text': 'hallway'
}
...
\end{Verbatim}
\end{MyColorPar}
\end{tiny}
\vspace{-4mm}
\caption{A typical exchange from a ParlAI training set involves messages passed using the observation/action dict (the test set would not include labels). Shown here is the bAbI dataset. 
\label{fig:typical}
}
\end{figure}

\subsection{Examples}

This directory contains examples of different mains:.

\begin{compactitem}
\item
{\tt\small display\_data:} display data from a particular task provided on the command-line.
\item
{\tt\small display\_model:} show the predictions of a provided model.
\item
{\tt\small eval\_model:} compute evaluation metrics for a given model on a given task.
\item
{\tt\small train\_model:} execute a standard training procedure with a given task and model, including logging and possibly alternating between training and validation.
\end{compactitem}

\vspace{2mm}
For example, one can display 10 random examples from the bAbI tasks \cite{weston2015towards}:\\
\textcolor{white}{~~~}{\tt\small python display\_data.py -t babi -n 10}
\vspace{2mm}\\
Display multitasking bAbI and SQuAD \cite{rajpurkar2016squad} at the same time:\\
\textcolor{white}{~~~}~~~{\tt\small python display\_data.py -t babi,squad} 
\vspace{2mm}\\
Evaluate an IR baseline model on the Movies Subreddit:\\
\textcolor{white}{~~~}~~~{\tt\small python eval\_model.py -m ir\_baseline -t `\#moviedd-reddit' -dt valid}
\vspace{2mm}\\
Train an attentive LSTM model on the SQuAD dataset with a batch size of 32 examples:\\
\textcolor{white}{~~~}~~~{\tt\small python train\_model.py -m drqa -t squad -b 32}

\subsection{Tasks}

\ifarxiv
Over 20 tasks are supported in the first release, including popular datasets such as 
SQuAD~\cite{rajpurkar2016squad}, bAbI tasks~\cite{weston2015towards},
QACNN and QADailyMail~\cite{hermann2015teaching},
CBT \cite{hill2015goldilocks},
bAbI Dialog tasks \cite{bordes2016learning},
Ubuntu \cite{lowe2015ubuntu} and 
VQA \cite{antol2015vqa}. 
\fi
All the datasets in the first release are shown in Fig.~\ref{fig:tasks}\footnote{All dataset descriptions and references are at \url{http://parl.ai} in the {\tt README.md} and {\tt task\_list.py}.}.


The tasks are separated into five categories:
\begin{compactitem}
\item Question answering (QA): one of the simplest forms of dialog, with only 1 turn per speaker. Any intelligent dialog agent should be capable of answering questions, and there are many kinds of questions (and hence datasets) that one can build, providing a set of very important tests. Question answering is particularly useful in that the evaluation is simpler than other forms of dialog if the dataset is labeled with QA pairs and the questions are mostly unambiguous.

\item Sentence Completion (Cloze Tests): the agent has to fill in a missing word in the next utterance in a dialog. Again, this is specialized dialog task, but it has the advantage that the datasets are cheap to make and evaluation is simple, which is why the community has built several such datasets. 
\item Goal-Oriented Dialog: a more realistic class of tasks is where there is a goal to be achieved by the end of the dialog. For example, a customer and a travel agent discussing a flight, one speaker recommending another a movie to watch, 
\if arxiv
two speakers agreeing when and where to eat together, 
\fi
and so on.
\item Chit-Chat: dialog tasks where there may not be an explicit goal, but more of a discussion --- for example two speakers discussing sports, movies or a mutual interest.
\item Visual Dialog: dialog is often grounded in physical objects in the world, so we also include dialog tasks with images as well as text. 
\if arxiv 
In the future we could also add other sensory information, such as audio.
\fi
\end{compactitem}

Choosing a task in ParlAI is as easy as specifying it on the command line,
as shown in the dataset display utility, Fig.~\ref{fig:example_output}.
If the dataset has not been used before, ParlAI will automatically download it. As all datasets are treated in the same way in ParlAI (with a single dialog API, see Sec.~\ref{sec:oad}),
a  dialog agent can switch training and testing between any of them. Importantly, one can specify many tasks at once (multitasking) by simply providing a comma-separated list, e.g.~the command line arguments {\tt\small -t babi,squad}, to use those two datasets, or even all the QA datasets at once ({\tt\small -t \#qa}) or indeed every task in ParlAI at once ({\tt\small -t \#all}). The aim is to make it easy to build and evaluate very rich dialog models.

Each task is contained in a folder with the following standardized files:

\begin{compactitem}
\item {\tt\small build.py}: file for setting up data for the task, including downloading the 
data the first time it is requested.
\item {\tt\small agents.py}: contains teacher class(es), agents
that live in the world of the task.
\item {\tt\small worlds.py}: optionally added for tasks that need to define new/complex environments.
\end{compactitem}

To add a new task, one must implement build.py to download any required data,
and agents.py for the teacher.
If the data consist of fixed logs/dialog scripts such as in many supervised datasets (SQuAD, Ubuntu, etc.) there is very little code to write. For more complex setups where an environment with interaction has to be defined,  new worlds and/or teachers can be implemented.

\subsection{Mechanical Turk}

An important part of ParlAI is seamless integration with Mechanical Turk for data collection, training or evaluation. 
Human Turkers are also viewed as agents in ParlAI and hence human-human, human-bot, or multiple humans and bots in group chat can all converse within the standard framework, switching out the roles as desired with no code changes to the agents. This is because Turkers also receive and send via the same interface: using the fields of the observation/action dict.
We provide two examples in the first release:

\begin{compactitem}
\item[(i)] {\bf qa\_collector:} an agent that talks to Turkers to collect question-answer pairs given a context paragraph to build a QA dataset, see Fig.~\ref{fig:mturk}.
\item[(ii)] {\bf model\_evaluator:} an agent which collects ratings from Turkers on the performance of a bot on a given task.
\end{compactitem}


Running a new MTurk task involves implementing and running a main file (like run.py) and defining
several task specific parameters for
the world and agent(s) you wish humans to talk to.
For data collection tasks the agent should pose the problem and ask the Turker for
e.g.~the answers to questions, see Fig.~\ref{fig:mturk}. Other parameters include
 the task description, the role of the Turker in the task, keywords to describe the task,
the number of hits and the rewards for the Turkers. 
One can run in a sandbox mode before launching the real task where Turkers are paid.

 For online training or evaluation, the Turker can talk to your machine learning agent, e.g.~LSTM, memory network or other implemented technique.
New tasks can be checked into the repository 
so researchers can share  data collection and data evaluation procedures and reproduce experiments.

\section{Demonstrative Experiment}\label{sec:exp}

To demonstrate ParlAI in action, we give results in Table~\ref{table:drqa-results} of DrQA, an attentive LSTM architecture with single task and multitask training on the SQuAD and bAbI tasks, a combination not shown before with any method, to our knowledge.

This experiment simultaneously shows the power of ParlAI --- how easy it is to set up this experiment --- and the limitations of current methods.
Almost all methods working well on
\ifarxiv
SQuAD have been designed to predict a phrase 
from the given context (they are given labeled start and end indices in training).
Hence, those models cannot be applied to all dialog datasets, e.g.~some of the bAbI tasks include yes/no questions, where yes and no do not appear in the context.  This highlights that
researchers should not focus models on a single dataset.
ParlAI does not provide start and end label indices
 as its API is dialog only, see Fig.~\ref{fig:oad}.
This is a deliberate choice that discourages such dataset overfitting/ specialization.
However, this also results in a slight drop in performance because less information is
given\footnote{As we now do not know the location of the true answer, we randomly pick the start and end indices of any context phrase matching the given training set answer, in some cases this is unique.} (66.4 EM vs. 69.5 EM, see \cite{drqa}, which is still in the range of many existing well-performing methods, see \url{https://stanford-qa.com}).

Overall, while DrQA can solve {\em some}
of the bAbI tasks and performs well on SQuAD,
it does not match the best performing methods on
bAbI \cite{seo2016query,henaff2016tracking}, and multitasking does not help. 
\fi
Hence, ParlAI lays out the challenge to the community to find
learning algorithms that are generally applicable and that benefit 
from training over many dialog datasets.

\section{Related Software}

There are many existing independent dialog datasets, and training code for individual models that work on some of them. Many are framed in slightly different ways (different formats, with different types of supervision), and ParlAI attempts to 
unify this fragmented landscape.

There are some existing software platforms that are related in their scope, but not in their specialization.
OpenAI's Gym and Universe\footnote{\tiny{\url{https://gym.openai.com/}} and \tiny{\url{https://universe.openai.com/}}}
are toolkits for developing and comparing reinforcement learning (RL) algorithms. Gym is for games like Pong or Go, and Universe is for online games and websites. Neither focuses on dialog or covers the case of supervised datasets as we do.

CommAI\footnote{\tiny \url{https://github.com/facebookresearch/CommAI-env}}
is a framework that uses textual communication for the goal of developing artificial general intelligence through incremental tasks that test increasingly 
\ifarxiv
more complex skills, as described in \cite{mikolov2015roadmap}.
\fi
CommAI is in a RL setting, and contains only synthetic datasets, rather than real natural language datasets as we do here. In that regard it has a different focus to ParlAI, which emphasizes the more immediate task of real dialog,
rather than directly on evaluation of machine intelligence. 

\begin{table}[t!]
  \begin{center}
    \resizebox{1\linewidth}{!}{
      {
    \begin{tabular}{rl|c|c}
     & Task &  Single   &  Multitask\\
\hline
bAbI 10k& 1: Single Supporting Fact & 100&	100\\
        & 2: Two Supporting Facts&	98.1&	54.3	\\
        & 3: Three Supporting Facts	&45.4&	58.1\\	
&4: Two Arg. Relations	&100	&100	\\
&5: Three Arg. Relations &98.9	&98.2\\	
&11: Basic Coreference	&100&	100	\\
&12: Conjunction	&100	&100	\\
&13: Compound Coref.&	100&	100	\\
&14: Time Reasoning&	99.8	&99.9	\\	
&16: Basic Induction	&47.7	&48.2\\
SQuAD & (Dev. Set)    	&66.4&	63.4\\	
\hline
    \end{tabular}
    }
    }
  \end{center}
    \caption{Test Accuracy of DrQA on bAbI 10k and SQuAD (Exact Match metric) using ParlAI. The subset of bAbI tasks for which the answer is exactly contained in the text is used.}
    \label{table:drqa-results}
\end{table}

\section{Conclusion and Outlook}

ParlAI is  a framework allowing the research community to
 share existing and new tasks for dialog 
as well as agents that learn on them,
and to collect and evaluate  conversations between agents and humans via Mechanical Turk.
We hope this tool enables systematic development and evaluation of dialog agents, helps push the state of the art in dialog
further, and benefits the field as a whole.

\ifarxiv
\section*{Acknowledgments}
We thank Mike Lewis, Denis Yarats, Douwe Kiela, Michael Auli, Y-Lan Boureau, Arthur Szlam, Marc'Aurelio Ranzato, Yuandong Tian, Maximilian Nickel, Martin Raison, Myle Ott,  Marco Baroni, Leon Bottou and other members of the FAIR team for discussions helpful to building ParlAI.
\fi

\small
\bibliography{emnlp2017}
\bibliographystyle{emnlp_natbib}

\end{document}